# Adaptive Transfer Learning in Deep Neural Networks: Wind Power Prediction using Knowledge Transfer from Region to Region and Between Different Task Domains

Aqsa Saeed Qureshi[1], and Asifullah Khan[*1,2]

[1]Department of Computer Science, Pakistan Institute of Engineering and Applied Sciences, Nilore-45650, Islamabad;

[2]Centre for Mathematical Sciences, Pakistan Institute of Engineering and Applied Sciences, Nilore-45650, Islamabad;

asif@pieas.edu.pk

**Abstract**

Transfer Learning (TL) in Deep Neural Networks is gaining importance because in most of the applications, the labeling of data is costly and time-consuming. Additionally, TL also provides an effective weight initialization strategy for Deep Neural Networks . This paper introduces the idea of Adaptive Transfer Learning in Deep Neural Networks (ATL-DNN) for wind power prediction. Specifically, we show in case of wind power prediction that adaptive TL of Deep Neural Networks system can be adaptively modified as regards training on a different wind farm is concerned. The proposed ATL-DNN technique is tested for short-term wind power prediction, where continuously arriving information has to be exploited. Adaptive TL not only helps in providing good weight initialization, but is also helpful to utilize the incoming data for effective learning. Additionally, the proposed ATL-DNN technique is shown to transfer knowledge between different task domains (wind power to wind speed prediction) and from one region to another region. The simulation results show that the proposed ATL-DNN technique achieves average values of 0.0637,0.0986, and 0.0984 for the Mean-Absolute-Error, Root-Mean-Squared-Error, and Standard-Deviation-Error, respectively.

**Keywords ----** Wind power prediction; Adaptive transfer learning; Deep learning; Ensemble learning

## 1. Introduction

Many countries across the world use wind power as a renewable energy resource. Accurate prediction of wind power plays a significant role in generating smooth power from a turbine. There are numerous factors which affect the predicted power of a wind power prediction system, like fluctuation in speed of the wind, geographical location, and climatic conditions. Short and long-term forecasting of wind are two main categories of wind power prediction strategies. When power prediction is carried out before minutes or a day, then it is called short-term forecasting of wind power. On the other hand, advance prediction of wind power before a year, months or even days is considered as long-term wind power prediction. For wind power forecasting, short-term based prediction of wind power is considered more reliable. Those countries, which are rich in wind farms mostly utilize different wind power prediction strategies. Wind farm prediction strategies are either based on statistical or physical models; moreover, hybrid models that are a combination of various physical and statistical models are also popular. Forecasting of wind power using physical methods depends on Numerical-Weather-Predictions like pressure, temperature, obstacles, the roughness of the surface, etc [1]. In case of statistical approaches, only historical data is considered without taking into consideration the meteorological conditions. Statistical methods use time-series and artificial intelligence based techniques to predict the wind power [2]. Hybrid approaches are a combination of statistical and physical methods that exploit both time-series and weather forecast techniques.

In the past various statistical approaches related to wind power, wind speed, and energy prediction have been proposed by various researchers. These statistical techniques include Radial Basis Function (RBF)[3], Neural Networks[4], Adaptive-Neuro-Fuzzy-Interface [5],Elman neural network[6], Auto-Regressive-Moving-Average (ARMA), Least-Square Support Vector Machine [7], adaptive wavelet neural network [8], extreme machine learning [9], deep learning [10] etc. In the same way, Landberg et al. [11]

suggested a wind power forecast system for wind farms that are connected to the electric grid. Cassola et al. [12] used the output from kalman filter as an analyzer in order to forecast wind speed and its corresponding energy. For reliable forecasting, Liu et al. [13] presented an interesting wind power prediction scheme, in which inputs are decomposed into frequency bands and then different models are trained according to SVM theory. Different kernels are used during the training of Liu's technique, whereas relative Mean-Squared-Error and Mean-Relative-Error are used as evaluation measures. In another wind power prediction strategy, Chitsaz et al. [14] introduced an effective wind power forecasting system that is constructed using Wavelet-Neural-Network (WNN). Haque et al.[15] proposed a probabilistic approach for wind power forecasting that comprised of a combination of quantile and deterministic regression approach. Similarly, in order to predict the output of wind power from wind farms, Mangalova et al. [16] suggested a wind power forecasting model in which combination of formal as well as heuristic methods are used. In Mangaloval's technique, during the construction of model, the basic approach used is K-nearest neighbor. Apart from statistical techniques, numerous hybrid approaches (that are a combination of statistical and physical methods) are also reported for reliable wind speed and power forecasting. A hybrid approach that uses mutual information (MI), evolutionary computing, wavelet transform, Particle-Swarm-Optimization (PSO), and adaptive-neuro-fuzzy system for prediction of wind power is developed by Osorio et al. [17].Similarly, a pattern sequence similarity based wind speed forecasting system is introduced by Bokde et al.[18]. Another methodology, that is based upon physical and statistical methods is presented by Peng et al.[19]. Similarly, Grassi et al. [20] used two hidden layers based Artificial Neural Network (ANN), for reliable forecasting of wind power. Hyperbolic-tangent and logarithmic-sigmoid are used as an activation function in the first and second layer, respectively, whereas the backpropagation algorithm is used for training. In another approach, a comparison between ARMA technique (an approach that uses five different ANN), and Adaptive-Neuro-Fuzzy-Interface-System based wind power forecasting is reported by Giorgi et al. [21] . Similarly, in order to improve the wind power forecasting, a novel spatio-temporal based approach is suggested by Xie et al. [22]. On the other hand, an evolutionary approach for optimizing wind turbine energy and its corresponding power factor is reported by Kusiak et al.[23]. Another interesting approach, which uses PSO and ANN for forecasting wind power is proposed by Amjady's et al. [24].

Ensemble-based regression techniques are also proposed by different researchers for robust and reliable forecasting of wind power. An ensemble based technique is proposed by Zameer et al. [25], whereby predictions from five different base-learners are provided to Genetic Programming based meta-learner. On the other hand, wind speed prediction strategy is developed by Hu et al. [26], that exploits the idea of Transfer Learning (TL). Hu's technique illustrates how old wind farm can be used as a source domain for TL and applies the gained knowledge into a newly built farm to reduce the training error. However, the idea of TL in combination with adaptive training and ensemble learning is not utilized to make the overall TL based technique an adaptive one. Mishra et al. [27] used RBF along with Legendre Neural Network for wind power forecast for the short-term interval. Recently, Deep Neural Network based wind power forecasting strategy (DNN-MRT) is developed by Qureshi et al.[28]. In DNN-MRT work, the concept of TL is used across wind farms situated in different areas of Europe. In previously proposed techniques, prediction by a single learner in most cases is not stable because of rapid fluctuations in the environmental conditions. Similarly, techniques in which different predictors are connected in series, the error usually goes on increasing as more and more predictors are added in series. Whereas, in ensemble-based wind power prediction strategies, the number of individual learners generally increases the generalization performance of meta-learner at a cost of computational complexity. In contrast, in the proposed ATL-DNN technique, the concept of TL is used not only within the wind farms but also across different wind farms. In one of our previous technique, "DNN-MRT", parameters of nine different base-learners in the source domain are optimized and used as a transferable knowledge for other wind farms. In comparison, in the proposed ATL-DNN technique, only three base-learners are adaptively trained after every four months period, and parameters are optimized using only on first base-learner (from source domain) and rest of the base-learners (both in source and target domains) are only fine-tuned. The proposed ATL-DNN technique provides good weight initialization for the training of base-learners and also inhibit the aggressive propagation of errors.

Moreover, the proposed ATL-DNN technique also show good results when pre-trained network on wind power related data from one region is adaptively fine-tuned for wind speed related task from different region. The contribution of the proposed work is as follows:

- An Adaptive wind power prediction system is proposed, whereby power prediction system for a different wind farm is generated through the combination of ensemble learning, adaptive training of Deep Neural Networks and TL.
- The ATL-DNN technique is shown to transfer knowledge between different task domains (wind power to wind speed prediction) and from one region to another region.
- Both inter and intra TL is used for the training of individual Deep Neural Networks based learners.

### 1.1. Research Work Related to Adaptive Learning in Deep Neural Network

In neural networks, weights are adaptively updated during the training phase. Specifically, in Deep Neural Networks, sometimes architecture becomes so deep that optimization of a lot of parameters becomes cumbersome during training. By using the idea of adaptive optimization of weights, different researchers used adaptive training of Deep Neural Networks in such a way that many machine learning tasks are performed just by adaptive tuning of pre-trained Deep Neural Networks. For example, for removal of noise from corrupted images, Agostinelli et al. [29] stacked denoising auto-encoder with an adaptive multi-column technique. To improve the performance of pre-trained denoising auto-encoder, Kim et al.[30] proposed adaptive learning strategies, in which already trained auto-encoder on noise-free spectra is stacked over the denoising auto-encoder trained on a different combination of spectra. During testing phase, fine-tuning is accomplished in such a way that stacked auto-encoder on the top is used to fine-tune the denoising auto-encoder at the bottom. Ochiai et al. [31] proposed adaptive training of different speakers using Deep Neural Network. In Ochiai's approach, a module that is dependent on the speaker is embedded in Deep Neural Network layer and during adaptive learning stage; only adaptation of speaker dependent module is performed. Another interesting adaptive learning approach across different language related task is proposed by Huang et al.[32]. In Huang's technique, hidden layers from Deep Neural Network trained on one type of language are used as a shared hidden layer across other Deep Neural Network for the recognition of another language. Using the idea of adaptive training in the proposed ATL-DNN technique, Deep Neural Network based adaptive online TL approach is introduced. In the proposed ATL-DNN technique, all of the base-learners are trained adaptively with time and thus can also be used to handle the online data that keeps on increasing with time. The rest of the paper is presented as such: section 2 is related to the proposed ATL-DNN technique while implementation details are discussed in section 3.Whereas, section 4 and section 5 are related to results and conclusion.

### 2. Proposed ATL-DNN Technique for Online Wind Power Prediction

In the proposed ATL-DNN technique, the input features of wind power are provided as input to the prediction system. The output is an hourly prediction of wind power. Figure 1 depicts the overall methodology of the proposed ATL-DNN technique showing input features and the corresponding output of the ensemble based adaptive learning system.

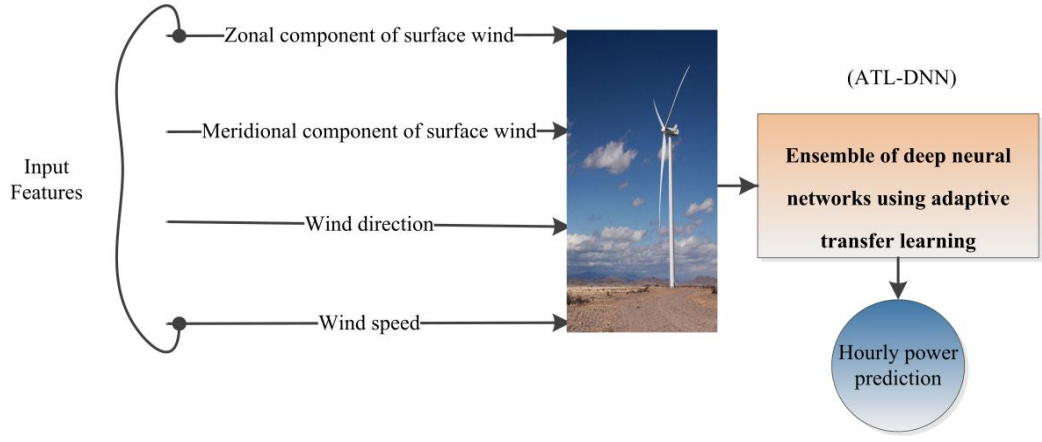

Figure 1: Proposed ATL-DNN technique with input features and predicted output

The Proposed ATL-DNN technique is an ensemble based regression approach, whereby, base-learners (auto-encoders) are trained adaptively after every four months. Figure 2 shows the basic idea of the adaptive TL. In ATL-DNN, out of the multiple wind farms data, auto-encoder uses first four months of a randomly selected $i^{th}$ wind farm data during the pre-training phase. After training, a wind farm on which a base- learner is pre-trained acts as a source domain for the online TL. For adaptive TL, other wind farms are used as a target domain, to adaptively fine-tune the pre-trained network (trained on $i^{th}$ wind farm). As more and more data is being generated continuously, that's why after every four months, the data previously generated is combined with newly generated data. The combined data is used to fine-tune the pre-trained base-learner to adaptively generate new base-learner. However, with time, the number of base- learners may increase, so only three latest trained auto-encoders are utilized as a base-learner in the ensemble learning.

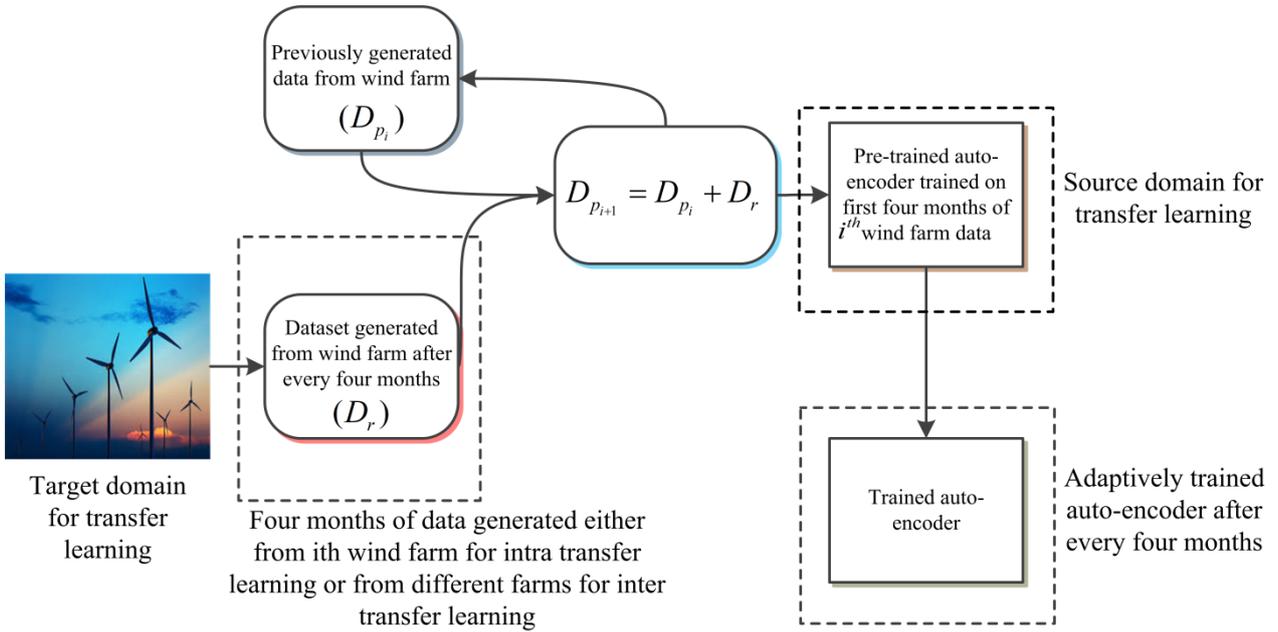

Figure 2: Use of adaptive TL for the adaptation of the wind power forecasting system

The ATL-DNN technique is basically an ensemble learning mechanism in which the concept of intra and inter TL is exploited for the training of the base-learners (within the same wind farm and also across different wind farms). As base-learners in ATL-DNN technique can be adaptively trained after every four months, that's why it can be used for online wind power forecasting. ATL-DNN works in two phases; during the first phase, training of first, second, and third auto-encoders is performed using first four, eight, and twelve months of wind farm data respectively. In the second phase, meta-learner (a DBN in our case) is trained on the predictions from trained auto-encoders, as well as the original features of the 13-16 months of wind farm data. In the end, the 17-20 months of data is used as a test data for assessing the performance of ATL-DNN. Flowchart of the proposed ATL-DNN technique is shown in Figure 3.

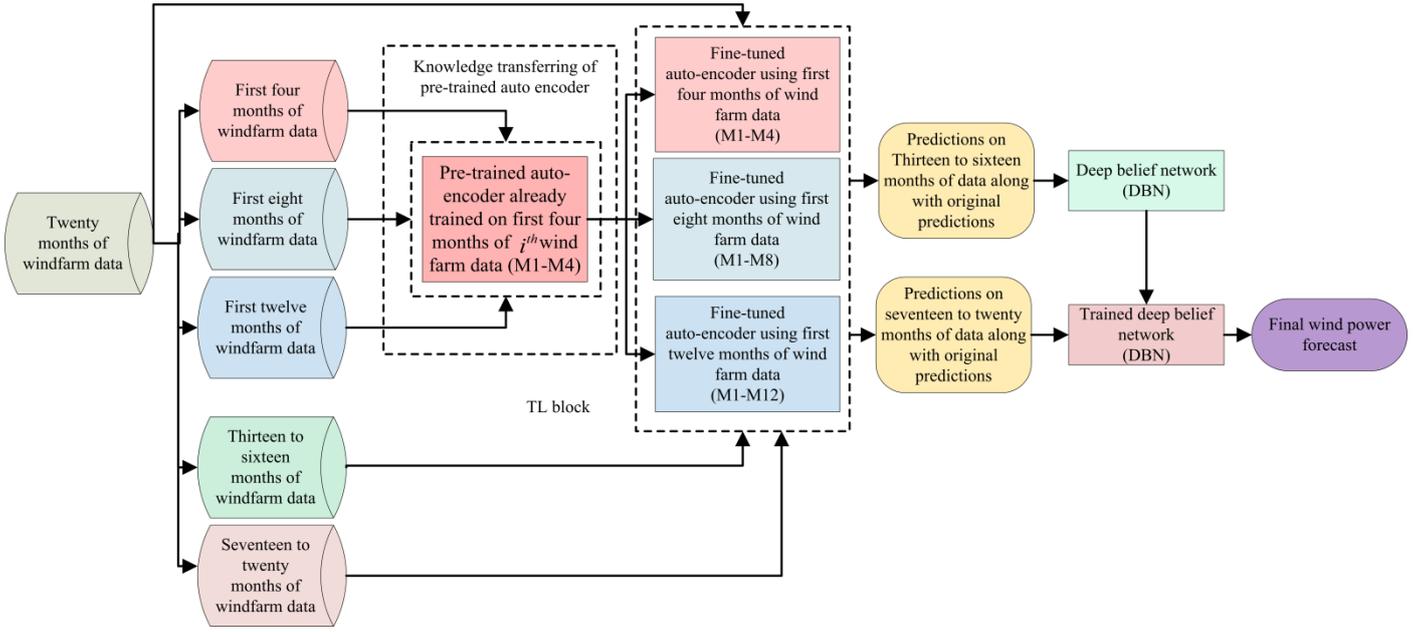

Figure 3: Flowchart of the proposed ATL-DNN technique. The adaptiveness upon the new arrival of data is achieved through TL block

### 2.1. Base and Meta-learner

In ATL-DNN technique, deep sparse auto-encoders and DBN are used for the construction of ensemble employing adaptive transfer learning based regression.

#### 2.1.1. Sparse Auto-encoder as Base-learner:

An auto-encoder is trained in such a way that it tries to copy its input data to its output. Its training phase comprises of the encoding and decoding phase. During the first phase, which is also called encoding phase, it tries to encode its input $x_{in}^{'}$ using the encoding function as shown in equation 1.

$$x_{in} = f(x_{in}^{'}) \qquad (1)$$

In the second phase, it decodes the encoded function ($x_{in}$) and reconstructs the original input $x_{in}^{'}$ as illustrated in equation 2.

$$r_{in} = f_r(x_{in}) \qquad (2)$$

During training, if an auto-encoder only tries to mimic the input data, this may result in overfitting. To avoid the overfitting, sparsity term is added during the training phase, which tends to enhance the generalization performance of the auto-encoder. Basically, an auto-encoder comprises of three layers: an input layer, hidden layer (encoded layer), and the last output layer (decoded layer). An auto-encoder is considered as under complete if the hidden layer has less number of neurons than the neurons in the input layer. For an auto-encoder during training, the use of least MSE as loss function may not be beneficial as it reduces the generalization performance [33]. However, the loss function used during the training of a sparse auto-encoder is given as under:

$$Loss = \frac{1}{N}\sum_{i=1}^{m}\sum_{j=1}^{n}(X_{ij} - \overline{X_{ij}})^2 + \lambda * \Omega_w + \beta * \Omega_s \qquad (3)$$

Where, $\frac{1}{N}\sum_{i=1}^{m}\sum_{j=1}^{n}(X_{ij} - \overline{X_{ij}})^2$ is the mean squared error, $\Omega_w$ is L2 weight regularization and mathematically expressed as:

$$\Omega_w = \frac{1}{2}\sum_{l}^{L}\sum_{i}^{m}\sum_{j}^{n}(W_{ij}^{l})^2 \qquad (4)$$

In equation 4, L and m represent the total number of hidden layers and examples respectively, whereas n is the number of variables used in the data. By adding L2 regularization as a regularization term in the loss function, will help in increasing the generalization property of the trained auto-encoder. $\lambda$ and $\beta$ are the coefficient associated with weight and sparsity regularization, respectively. $\Omega_s$ is the sparsity regularization term and can be expressed mathematically as:

$$\Omega_s = \sum_{i=1}^{s} KL_D(\frac{p}{p_i}) = \sum_{i=1}^{s} p\log(\frac{p}{p_i}) + (1-p)\log(\frac{1-p}{1-p_i}) \qquad (5)$$

The sparsity regularization enforces the sparsity constraint on the output layer. $p$ is the desired activation value of a neuron i and $p_i$ is the average activation value of a neuron. Whenever the actual and desired activation value of $i^{th}$ neuron are same, then the sparsity regularization term will be zero. Whereas, with the increase in the difference between $p$ and $p_i$, $\Omega_s$ term will increase. Weight and sparsity regularization terms in loss function help in adjusting the weights during the training of an auto-encoder, which ultimately increases the generalization performance.

Another way to boost the performance of an auto-encoder is to train it as an over-complete auto-encoder, such that the number of neurons in the hidden layer is greater than the input features. Sparsity term in loss function helps in learning some useful properties during the training phase, instead of only copying the input data. Mostly, an auto-encoder is trained by using a single hidden layer for encoding, whereas output layer is used to decode the encoded input. Auto-encoders can be made deeper, which offers many useful properties in comparison to single layer auto-encoder.

In general, two steps are involved in the training of deep, sparse auto-encoders. In the first step, the output layer of sparse auto-encoder is removed after pre-training and then the output from the hidden layer is provided as an input to the second sparse auto-encoder. Second auto-encoder is trained in the same way as the first one; the process goes on until the desired number of auto-encoders has been trained. After pre-training, all the auto-encoders are stacked to form a feed-forward neural network. Pre-training may also help good initial weights for the Feed Forward Neural Network (FFNN). Figure 4 shows the individual training of sparse auto-encoder. Below is the pseudo code used for the individual training of a sparse auto-encoder during the pre-training stage.

$Algorithm:$

$Pre-training\ of\ Sparse\ Auto-encoder\ (SAE)$

$Procedure\ (E, B, x_{in}', \alpha, r_{in}, \Omega_w, \lambda, \Omega_s, \beta)$

$E\ (number\ of\ training\ epochs)$

$B\ (numbe\ of\ Batches)$

$x_{in}'\ (input)$

$r_{in}\ (reconstructed\ input)$

$\Omega_w\ (L_2\ weight\ regularization\ term)$

$\lambda\ (Coefficient\ of\ L_2\ weight\ regularization\ term)$

$\Omega_s\ (Sparsity\ regularization\ term)$

$\beta\ (Coefficient\ of\ sparsity\ regularization\ term)$

$m\ (number\ of\ examples)$

$n\ (number\ of\ variables)$

$\frac{1}{N}\sum_{i=1}^{m}\sum_{j=1}^{n}(X_{ij}-\overline{X_{ij}})^2\ (Mean\ Squared\ Error)$

$for\ 0 \to E$

$\quad for\ 0 \to B$

$\quad\quad x_{in} = f(x_{in}')$

$\quad\quad r_{in} = f_r(x_{in})$

$\quad\quad Loss = \frac{1}{N}\sum_{i=1}^{m}\sum_{j=1}^{n}(X_{ij}-\overline{X_{ij}})^2 + \lambda*\Omega_w + \beta*\Omega_s$

$\quad\quad Compute\ gradiant\ of\ Loss\ function\ with\ respect\ to\ weights$

$\quad\quad W_{new} = W_{old} - \alpha*gradiant$

$\quad end\ for$

$end\ for$

$end\ precodure$

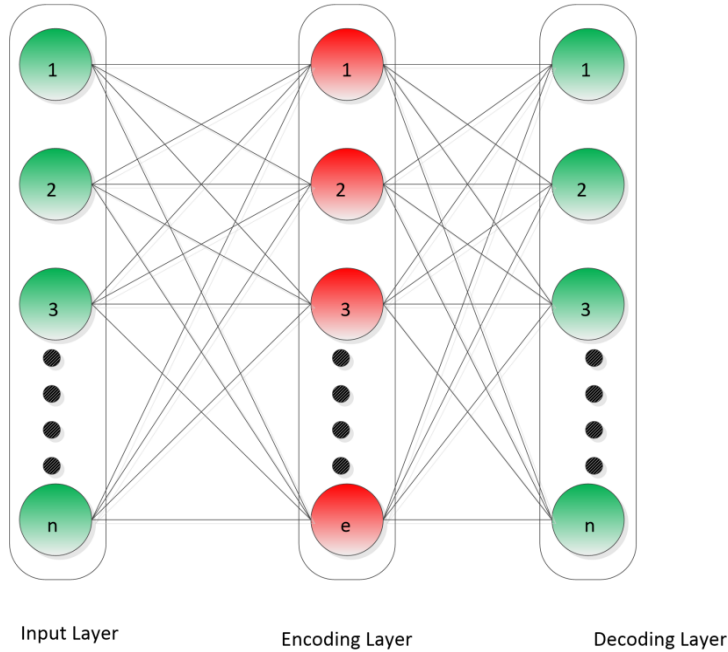

Figure 4: Pre-training of an auto-encoder

After the pre-training, the next step is to fine-tune the network. Figure 5 shows the stacking of a required N number of pre-trained auto-encoders, in the form of FFNN. This stacked network is then fine-tuned using Backpropagation algorithm, whereas weights are initialized as obtained in the pre-training phase. According to the universal approximation theorem, a single layer auto-encoder can learn any function, provided a sufficient number of hidden layer neurons, but its disadvantage is that it increases the computational cost [34]. Another way to reduce computational cost is to increase the depth of the network because the increase in depth may exponentially reduce the computational power and also may require fewer training examples [35].

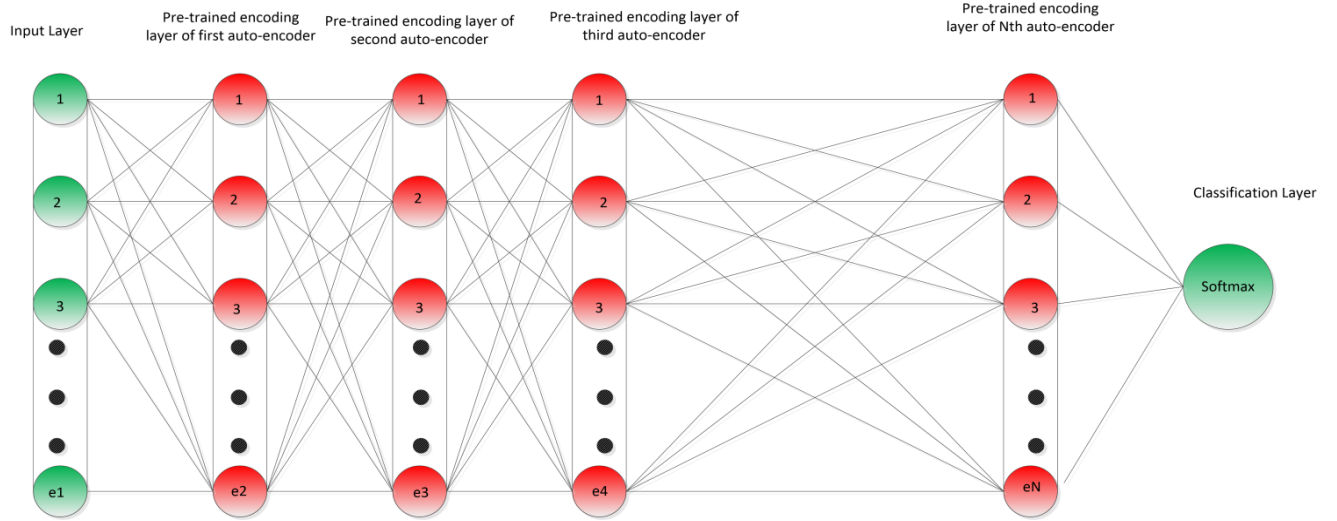

Figure 5: Stacking of N number of pre-trained auto-encoders, in the form of FFNN

### 2.1.2. Deep Belief Network Based Meta-Learner

DBN [36] is considered as a generative model, in which Restricted Boltzmann Machine (RBM) [37] is used as a basic building block. RBM is originated from Boltzmann Machine (BM). In RBM unlike BM, there is no connection within the same layer. A single unit of RBM is comprised of visible ($v'$) and hidden ($h'$) layers, the joint energy of hidden and visible layer is calculated using equation 6. RBM, which is the basic unit of DBN, is shown in Figure 6.

$$E_j(v',h') = -\sum_{i=1}^{m} a_i'v_i' - \sum_{j=1}^{n} b_j'h_j' - \sum_{j=1}^{n}\sum_{i=1}^{m} w_{ji}'v_i'h_j' \qquad (6)$$

Joint probability between visible and hidden layer is expressed in equation 7:

$$P(v',h') = \frac{e^{-E_j(v',h')}}{Z_{E_j}} \qquad (7)$$

Whereas, $Z_{E_j}$ is the partition function expressed by adding all the possible energy configuration values between hidden and visible layer as given below:

$$Z_{E_j} = \sum_{v',h'} e^{-E_j(v',h')} \qquad (8)$$

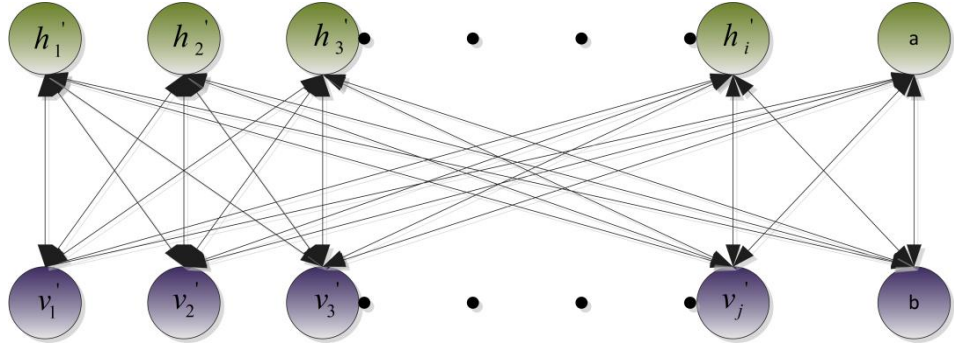

Figure 6: Restricted Boltzmann Machine (RBM)

Unfortunately, the partition function is intractable to compute; therefore it is difficult to use joint probability as loss function. Another alternative is to use Gibb's sampling theorem for updating the weights during the RBM's training. According to Contrastive Divergence (CD-k) theorem, state of hidden layer ($h'$) neurons can be computed from the visible layer ($v'$). Equation 9 shows the computation of hidden layer state using visible layer.

$$p(h_i'=1/v') = func'(\sum_{j=1}^{m} v_j' w_{ji}' + b_i') \tag{9}$$

In a similar manner, the state of the visible layer can be computed using equation 10.

$$p(v_j'=1/h') = func'(\sum_{i=1}^{n} h_i' w_{ji}' + a_j') \tag{10}$$

Alternatively hidden or visible layer state is computed k number of times in order to reconstruct visible units from hidden units. Weights and biases are updated using equations 11, 12, and 13, whereby, $\eta$ is the learning rate.

$$\nabla w_{ji}' = \eta((v_i' h_j')_{t_0} - (v_i' h_j')_{t_k}) \tag{11}$$

$$\nabla b_j' = \eta((h_j')_{t_0} - (h_j')_{t_k}) \tag{12}$$

$$\nabla \alpha_i' = \eta((v_i)_{t_0} - (v_i)_{t_k}) \tag{13}$$

In case of DBN, first greedy layer-wise pre-training of RBM is performed in such a way that output from first RBM is considered as an input for the second RBM. After that, different RBMs are stacked over one another to form a network. In the second step, fine-tuning of RBM is performed using Backpropagation algorithm [38] or, in case of generative fine-tuning, the wake-sleep algorithm is used.

In ATL-DNN, DBN based meta-learner is trained on 13-16 months of wind farm data. In the first step, predictions of the base-learners on 13-16 months of data is taken from the base- learners. Then the predictions along with the original features are used as an input for DBN. After training of the DBN, test data (which is comprised of 17-20 months of data) is used to evaluate the working of ATL-DNN.

### 2.2. Dataset and Feature Extraction

The dataset used is obtained from European-Center of Medium-range Weather-Forecasts (ECMWF). The dataset is comprised of three years of data from five different wind farms located near Europe, but only twenty months of data is used in proposed ATL-DNN technique. Power measurement against each wind farm has a temporal resolution of one hour, moreover, to ensure the scale-free comparison, power measurements are normalized between 0 and 1. Beside each power measurement, Zonal ($Z_s$) and meridional component ($M_s$) of surface wind at the height of about 10 m above the ground along with corresponding wind direction ($D_w$) and speed ($S_w$) are considered as a weather forecast. Considering power measurement at time t, weather forecasts are released at time t-12, t-24, t-36 and t-48 hours. To select the useful feature set from the four different sets of features, MI-based feature selection is used. MI selects feature that has the highest value of MI with respect to actual forecasted power; all of the remaining feature sets are discarded.

As a prediction of power at time t may depend on the previously generated power and the associated feature set, that's why predicted power of last 24 hours are combined with the MI based selected features to form a new feature set. Mathematically, the feature set of power at time t can be expressed as:

$$P_w(t) = \begin{bmatrix} P_w(t-1), P_w(t-2), P_w(t-3)\ldots\ldots\ldots, P_w(t-24) \\ D_w(t), D_w(t-1), D_w(t-2), \ldots, D_w(t-24), \\ Z_s(t), Z_s(t-1), Z_s(t-2), \ldots, Z_s(t-24), \\ M_s(t), M_s(t-1), M_s(t-2), \ldots M_s(t-24), \\ S_w(t), S_w(t-1), S_w(t-2), \ldots S_w(t-24) \end{bmatrix} \qquad (14)$$

### 2.3. Dataset Processing for Adaptive Training

In the proposed ATL-DNN technique, five different wind farm data located near Europe are used. Now, since in ATL-DNN technique, base-learners (auto-encoders) are trained adaptively with respect to time, that's why first four months of data (DS3) is used for the training of the first auto-encoder. However, the second auto-encoder is trained by using first eight months of wind farm data (DS4), while the third auto-encoder is trained on the first twelve months of data (DS5). After training of the base-learners, predictions on 13-16 months of wind farm data (DS6) along with the original features are used to tune DBN. To check the performance of the proposed ATL-DNN technique, 17-20 months of data (DS7) is reserved as a test set. This test set is then provided to the trained ensemble based model. Flowchart of dataset division is shown in Figure 7.

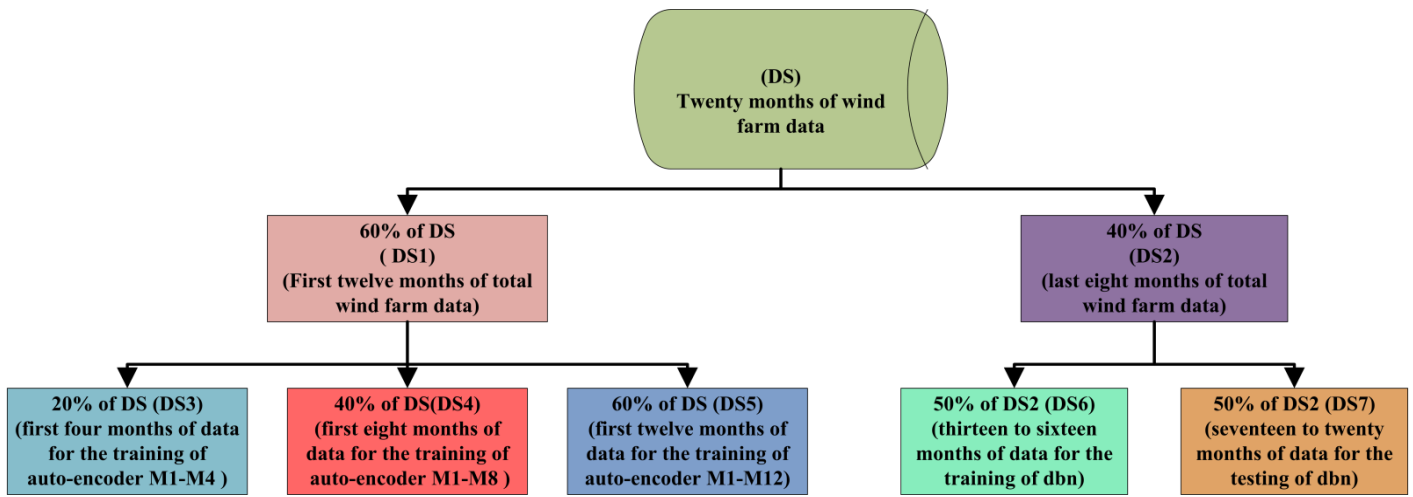

Figure 7: Dataset division

## 2.4. Training of Base-learners Using Inter and Intra Adaptive TL

The proposed ATL-DNN technique is an ensemble-based approach in which three base-learners (auto-encoders) are trained adaptively after every four months. Moreover, TL across the base-learners, during training of all wind farm datasets, is classified either as intra TL or inter TL.

### 2.4.1. Intra Adaptive TL within the $i^{th}$ Wind Farm

For a randomly selected $i^{th}$ wind farm, intra TL is used. In intra TL, three base-learners are trained adaptively as more and more wind farm data is added after every four months. First, a base-learner is trained using first four months of $i^{th}$ wind farm dataset and is considered as the pre-trained auto-encoder. After training of first base-learner, a second base-learner is formed by only fine-tuning the pre-trained auto-encoder using eight months of data. In a similar way, a third base-learner is formed by fine-tuning of the pre-trained auto-encoder using first twelve months of data. Adaptive training of the base-learners after every four months within the $i^{th}$ wind farm data is considered as intra TL. In intra TL, there is no need to train all of the three base-learners from scratch, only first base-learner is trained from scratch to develop a pre-trained auto-encoder, while the remaining two base-learners are formed by only fine-tuning the pre-trained auto-encoder. Intra TL requires that the source and target domain for TL belong to the same wind farm. Figure 8 demonstrates the idea of intra TL used in ATL-DNN.

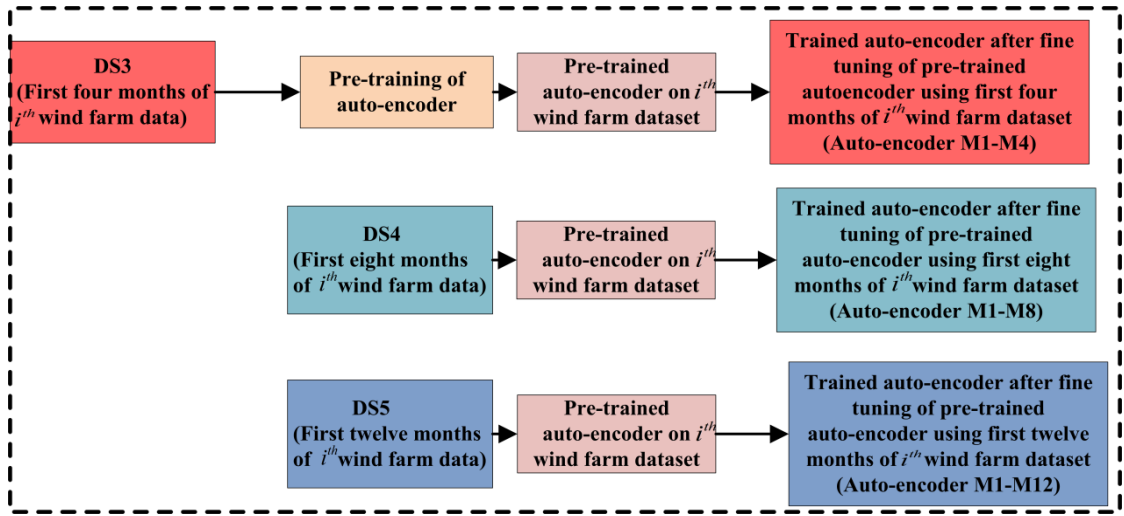

Figure 8: Intra TL within wind farm 2

### 2.4.2. Inter Adaptive TL Across Wind Farms

After the training of base-learners using intra TL for $i^{th}$ wind farm data, the next phase is to train base-learners for remaining wind farms using the inter TL approach. During the training of remaining base-learners, pre-trained auto-encoder using first four months of $i^{th}$ wind farm dataset is fine-tuned adaptively using first four, eight, and twelve months of data from remaining wind farm datasets. In the current work, wind farm2 is set as $i^{th}$ wind farm, and the remaining four wind farms are used as a target domain for inter TL. The concept of inter TL across wind farm datasets is depicted in Figure 9. By introducing the concept of intra and inter TL in the proposed ATL-DNN technique, parameters of auto-encoder are optimized using only first four months of $i^{th}$ wind farm dataset. All of the remaining fourteen base-learners (three for each wind farm) are only fine-tuned, and so there is no need to train base-learners from scratch even for different wind farms.

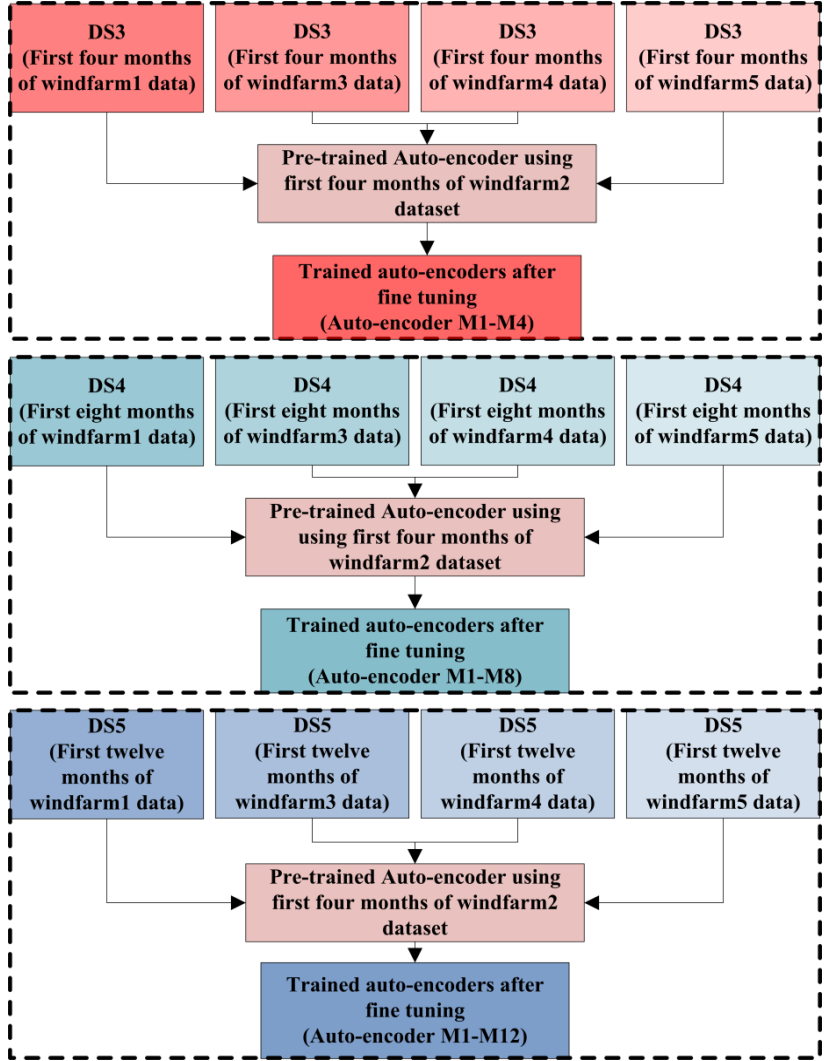

Figure 9: Inter TL across different wind farms

### 2.5. Exploitation of Intra or Inter TL for the Adaptive Training in ATL-DNN

TL is a machine learning technique that can increase the generalization of a trained network across different domains. In literature TL is used by many researcher for increasing the performance of different machine learning related tasks [39] [40][41][42][43]. In TL, there are basically two tasks, first one is source task, while the second one is target task. The first step of TL is to gain knowledge of the source task either by extracting some useful features or by transferring the pre-trained network that is trained on the source task. Target task uses the information transferred by source task. If TL helps in improving the performance of a target task, then it is useful and referred to as positive TL, otherwise, it results in negative TL. When the source task is somehow related to the target domain task, then it will result in positive TL. Otherwise, use of TL may also have a negative impact on target domain performance. In machine learning related applications, TL is mostly used for two purposes:

i. Performance of a classifier or learner depends on the amount of training data. Large amount of training data helps in increasing the generalization performance of the classifier. In most of the cases, sufficient amount of data is not available for the training of the network. In such a scenario, TL is helpful in gaining knowledge from the source domain that has sufficient amount of training data and then, applying the gained knowledge to the target domain.

ii. If a large amount of training data is available in the target domain, but training is taking sufficient time, then TL is helpful in this case as well. In such a scenario by using a pre-trained network from a source domain can only fine-tune it on the target domain to achieve appreciable performance as if the training of the target domain is performed from scratch.

The main idea proposed in this work; adaptive TL, is a type of TL in which adaptive training of the network is performed across related machine learning tasks. In ATL-DNN technique, adaptive TL is applied on five wind farms from different regions. ATL-DNN technique is based on the ensemble of Deep Neural Network, in which adaptive TL is applied across the base-learners of five wind farms datasets. Pre-trained auto-encoder (as first base-learner) from the randomly selected $i^{th}$ wind farm is considered as source domain for TL. When fine-tuning of pre-trained based-learners (from an $i^{th}$ wind farm) is fine-tuned adaptively to form base-learners for an $i^{th}$ wind farm (as target domain) then it is considered as intra TL. However, when the base-learners are formed adaptively after fine-tuning of pre-trained base-learner (source domain) for wind farms (except $i^{th}$ wind farm) then it is considered as an inter TL.

## 3. Implementation Details

The simulations are carried out on a desktop personal computer having 16 GB RAM, core (TM) i7-33770, and 64-bit operating system with 3.46 GHz processor. Windows 7 is used as an operating system and Matlab 2015(b) as a programming environment. For the implementation of deep sparse auto-encoder, NN-toolbox is used, whereas, the functions used during the training are mentioned in Table 1. For the implementation of meta-learner, DBN-Toolbox is downloaded from GitHub [44].

Table 1: Implementation detail related to the implementation of deep, sparse auto-encoder

| Function name | Purpose |
|---|---|
| **trainAutoencoder(X)** | Return a trained auto-encoder whereas X is the training data. |
| **stack()** | Used to stack the pre-trained auto-encoders |
| **train()** | The function is used to fine-tune the train stacked pre-trained auto-encoders |

### 3.1. Parameters Setting of the Base-Learners

In ATL-DNN technique, inter and intra TL overcomes the requirement of adjusting parameters for every base-learner. As a parameter tuning of only single auto-encoder is required, therefore, it not only reduces the training time but, also results in effective performance. Table 2 shows the details of the parameters of trained auto-encoder.

Table 2: Parameters of trained auto-encoder

| Layer | No of Neurons | Maximum Epoch | L2 Weight Regularization | Sparsity Regularization | Sparsity Proportion |
|---|---|---|---|---|---|
| 1 | 500 | 500 | 0.00003 | 4 | 0.15 |
| 2 | 400 | 250 | 0.00001 | 4 | 0.1 |
| 3 | 350 | 200 | 0.00001 | 4 | 0.1 |
| 4 | 300 | 200 | 0.00001 | 4 | 0.1 |
| 5 | 250 | 150 | 0.00001 | 4 | 0.1 |

### 3.2. Parameter Settings of Meta-learner

In the proposed ATL-DNN technique, DBN is used as a meta-learner, which is trained on 13-17 months of wind farm data. Table 3 shows the parameter setting for wind farm 1 and 2. Whereas, Table 4 illustrates the parameters of DBN setting for wind farm3, wind farm4, and wind farm5.

Table 3: Parameters of trained DBN for wind farm1 and wind farm2

| Parameter | Value |
|---|---|
| Number of neurons in layer 1 | 120 |
| Number of neurons in layer 2 | 50 |
| Number of neurons in layer 3 | 20 |
| Number of neurons in layer 4 | 05 |
| Number of epochs | 300 |
| Batch size | 10 |
| Momentum | 0.01 |
| Learning rate | 0.001 |

Table 4: Parameters of trained DBN for wind farm3, wind farm4, and wind farm5

| Parameter | Value |
|---|---|
| Number of neurons in layer 1 | 545 |
| Number of neurons in layer 2 | 300 |
| Number of neurons in layer 3 | 250 |
| Number of neurons in layer 4 | 50 |
| Number of neurons in layer 5 | 20 |
| Number of neurons in layer 6 | 2 |
| Number of epochs | 300 |
| Batch size | 10 |
| Momentum | 0.01 |
| Learning rate | 0.001 |

### 3.3. Performance Evaluation Measures

In order to check the performance of ATL-DNN technique, RMSE, MAE, and SDE are used as evaluation measures. Let's assume $p_{original}$ is the desired wind power, whereas $p_{predict}$ is the predicted power. If, and $error_{mean}$ is the mean value of the error, then, all the three evaluation measures can be calculated as:

$$RMSE = \sqrt{\frac{1}{m}(\sum_{i=1}^{m} p_{original} - p_{predict})^2} \qquad (15)$$

$$MAE = \frac{1}{m}\sum_{i=1}^{m} \| p_{original} - p_{predict} \| \tag{16}$$

$$SDE = \sqrt{\frac{1}{m}\sum_{i=1}^{m}(error - error_{mean})^2} \tag{17}$$

## 4. Results

In the proposed ATL-DNN technique, auto-encoder trained on the first four months of data (from wind farm2) is used as a source of transferring knowledge for intra and inter TL. Moreover, parameters are optimized on the basis of RMSE, and test data is used to evaluate the performance of the trained ATL-DNN. Table 5. shows the performance of the conventional Machine Learning (ML) techniques such as Support Vector Regressor (SVR) and Autoregressive Integrated Moving Average (ARIMA)) on the wind farm dataset.

Table 5: Performance of the conventional ML techniques

|  | ARIMA | | | SVR (linear kernel) | | | SVR (rbf kernel) | | |
|---|---|---|---|---|---|---|---|---|---|
|  | MAE | SDE | RMSE | MAE | SDE | RMSE | MAE | SDE | RMSE |
| Wind farm1 | 0.4470 | 0.3776 | 0.5410 | 0.0722 | 0.0982 | 0.0988 | 0.2319 | 0.2710 | 0.2838 |
| Wind farm2 | 0.4432 | 0.3956 | 0.5543 | 0.0688 | 0.1040 | 0.1040 | 0.2463 | 0.2894 | 0.2941 |
| Wind farm3 | 0.5711 | 0.4929 | 0.6832 | 0.1019 | 0.1380 | 0.1384 | 0.3047 | 0.3415 | 0.3520 |
| Wind farm4 | 0.4897 | 0.4433 | 0.6117 | 0.1234 | 0.1471 | 0.1479 | 0.2724 | 0.3100 | 0.3101 |
| Wind farm5 | 0.5071 | 0.4589 | 0.6178 | 0.0943 | 0.1275 | 0.1282 | 0.2906 | 0.3354 | 0.3377 |

### 4.1. Comparative Evaluation of the Base and Meta learners on Test Set Using Ten Independent Runs

In the ensemble based ATL-DNN technique, three auto-encoders as base-learners are trained. After training of the base-learners, a meta-learner, which is DBN in this case, provides the final prediction using the individual predictions of the base learners. During training, parameters are optimized according to RMSE, but final results are reported not only on RMSE but also on MAE and SDE.

After averaging the performance of ten independent runs, the performance of the base and meta learners is reported in terms of RMSE, MAE, and SDE in Table 6, Table 7, and Table 8, respectively. The RMSE of base and meta learner for different wind farms is shown in Table 6. Table 6 depicts that performance of the meta-learner is better in comparison to each of the individual base-learners. In the proposed ATL-DNN technique, during the training, optimization of parameters is performed by evaluating the RMSE. However, Table 7 and 8 show that the performance of the proposed ATL-DNN is not only good in terms of RMSE (on the basis of which parameters are optimized) but also in terms of MAE and SDE. In summary, Tables [6-8] show that final wind power forecast by the meta-learner is better and has less variation in comparison to the individual base-learners, which indicates that the ATL-DNN technique is more effective and robust.

Table 6: Performance of the proposed ATL-DNN technique in terms of RMSE

|  | Auto-encoder M1-M4 (RMSE) | Auto-encoder M1-M8 (RMSE) | Auto-encoder M1-M12 (RMSE) | DBN(RMSE) |
|---|---|---|---|---|
| **Wind farm1** | 0.0924 ± 0.0055 | 0.0826 ± 0.0015 | 0.0819 ± 0.0015 | 0.0791 ± 0.0003 |
| **Wind farm2** | 0.1134 ± 0.0058 | 0.1000 ± 0.0029 | 0.0957 ± 0.0015 | 0.0939 ± 0.0003 |
| **Wind farm3** | 0.1260 ± 0.0058 | 0.1104 ± 0.0016 | 0.1082 ± 0.0015 | 0.1080 ± 0.0004 |
| **Wind farm4** | 0.1197 ± 0.0072 | 0.1075 ± 0.0010 | 0.1065 ± 0.0022 | 0.1013 ± 0.0003 |
| **Wind farm5** | 0.1231 ± 0.0030 | 0.1146 ± 0.0019 | 0.1118 ± 0.0010 | 0.1109 ± 0.0003 |

Table 7: Performance of the proposed ATL-DNN technique in terms of MAE

|  | Auto-encoder M1-M4 (MAE) | Auto-encoder M1-M8 (MAE) | Auto-encoder M1-M12 (MAE) | DBN (MAE) |
|---|---|---|---|---|
| **Wind farm1** | 0.0646 ± 0.0033 | 0.0559 ± 0.0013 | 0.0558 ± 0.0017 | 0.0551 ± 0.0004 |
| **Wind farm2** | 0.0757 ± 0.004 | 0.0640 ± 0.0027 | 0.0610 ± 0.0014 | 0.0594 ± 0.0003 |
| **Wind farm3** | 0.0821 ± 0.0045 | 0.0707 ± 0.0012 | 0.0691 ± 0.0016 | 0.0698 ± 0.0004 |
| **Wind farm4** | 0.0809 ± 0.0061 | 0.0697 ± 0.0009 | 0.0691 ± 0.0020 | 0.0674 ± 0.0003 |
| **Wind farm5** | 0.0788 ± 0.0022 | 0.0699 ± 0.0017 | 0.0680 ± 0.0011 | 0.0671 ± 0.0002 |

Table 8: Performance of the proposed ATL-DNN technique in terms of SDE

|  | Auto-encoder M1-M4 (SDE) | Auto-encoder M1-M8 (SDE) | Auto-encoder M1-M12 (SDE) | DBN (SDE) |
|---|---|---|---|---|
| **Wind farm1** | 0.0923 ± 0.0054 | 0.0826 ± 0.0015 | 0.0819 ± 0.0015 | 0.0789 ± 0.0003 |
| **Wind farm2** | 0.1121 ± 0.0058 | 0.0994 ± 0.0028 | 0.0957 ± 0.0015 | 0.0939 ± 0.0003 |
| **Wind farm3** | 0.1259 ± 0.0058 | 0.1104 ± 0.0016 | 0.1082 ± 0.0014 | 0.1078 ± 0.0004 |
| **Wind farm4** | 0.1196 ± 0.0071 | 0.1075 ± 0.0010 | 0.1064 ± 0.0022 | 0.1008 ± 0.0002 |
| **Wind farm5** | 0.1213 ± 0.0029 | 0.1144 ± 0.0019 | 0.1117 ± 0.0010 | 0.1108 ± 0.0003 |

Figure10 shows the performance of the three base-learners and the corresponding meta-learner. The low value of RMSE shows that the concept of intra TL helps in achieving good performance. Figure 11, on the other hand, shows the performance on wind farm1 and it is clearly shown that the TL is effective and reliable when wind farm1 is used as a target domain (for inter TL). Figure 12 shows the performance of wind farm3; here too the concept of TL is effective and it can be observed that the performance of the meta-learner is robust in comparison to the individual base-learners. Similarly, Figure 13 and 14 show performance on wind farm4 and wind farm5 respectively, which indicate that the idea of inter TL is successful as almost all of the evaluation metrics have low values. This idea of inter TL has another advantage of saving precious time required to develop fully trained base-learners from scratch.

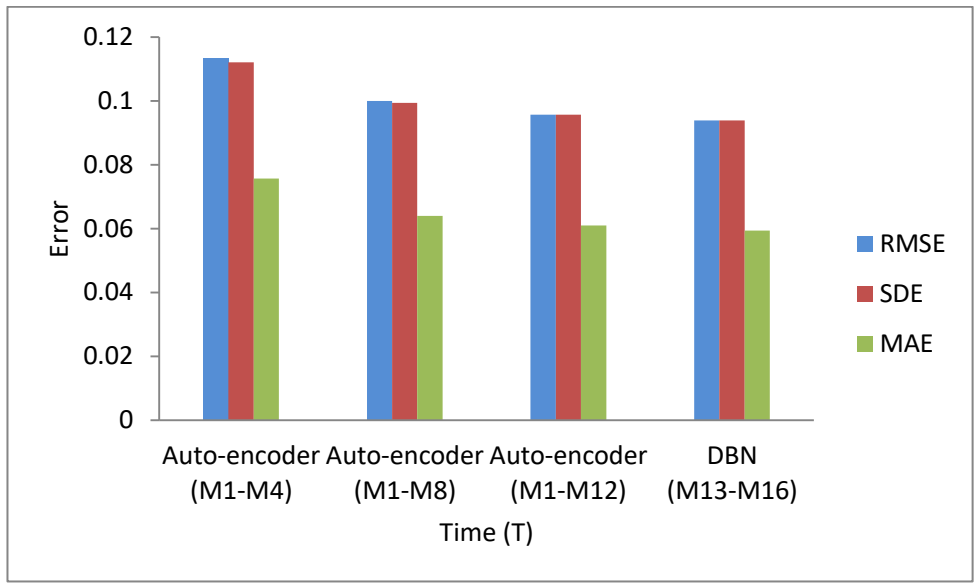

Figure 10: Performance of the proposed ATL-DNN technique on wind farm2

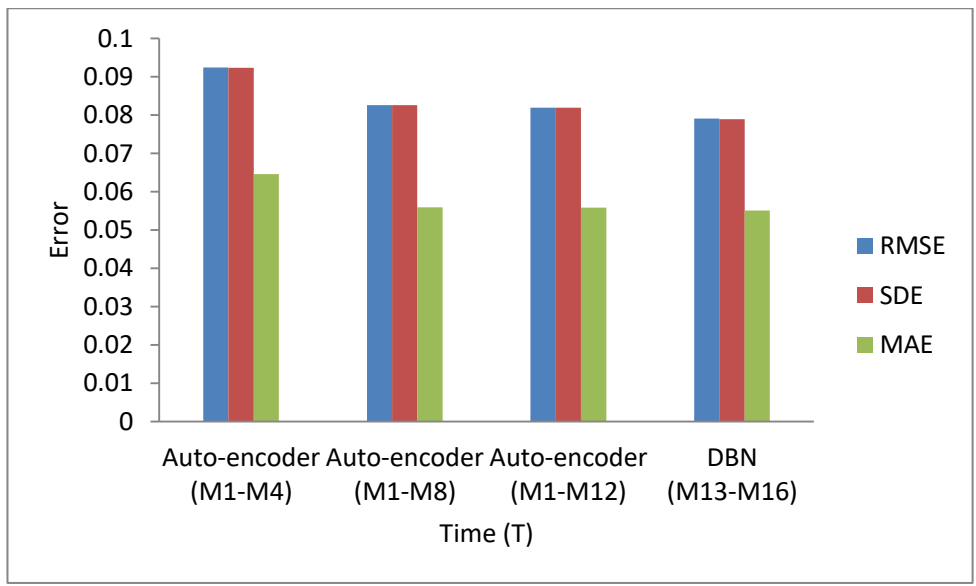

Figure 11: Performance of the proposed ATL-DNN technique on wind farm1

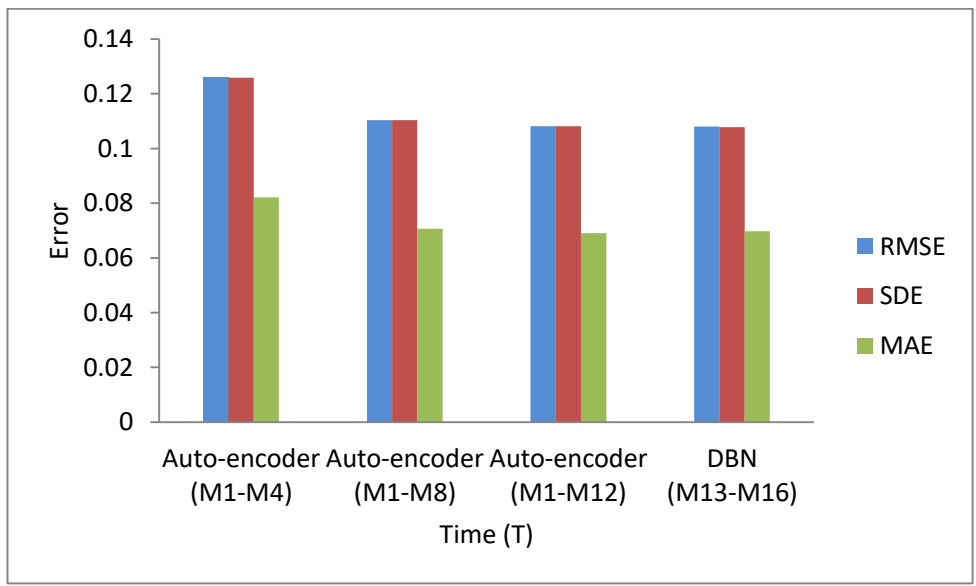

Figure 12: Performance of the proposed ATL-DNN technique on wind farm3

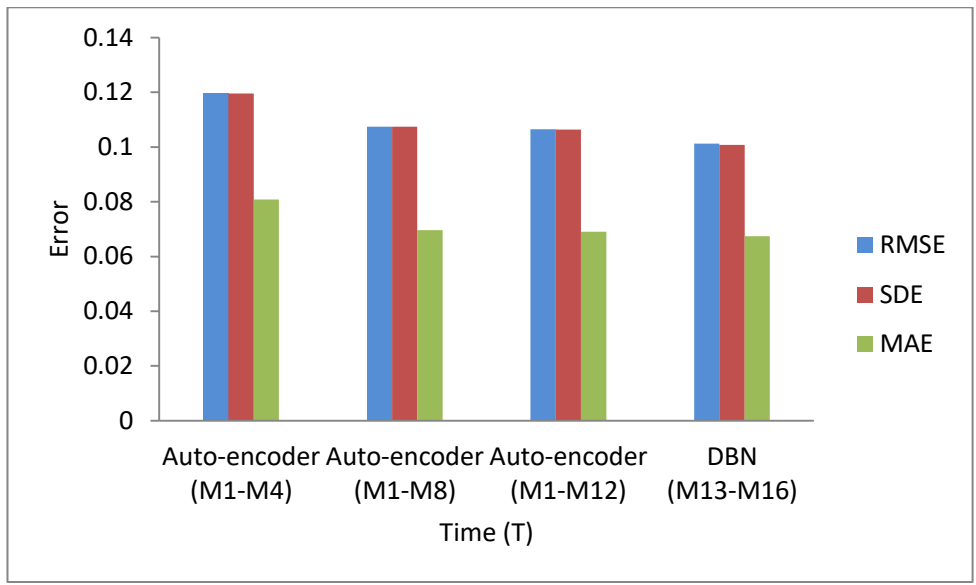

Figure 13: Performance of the proposed ATL-DNN technique on wind farm4

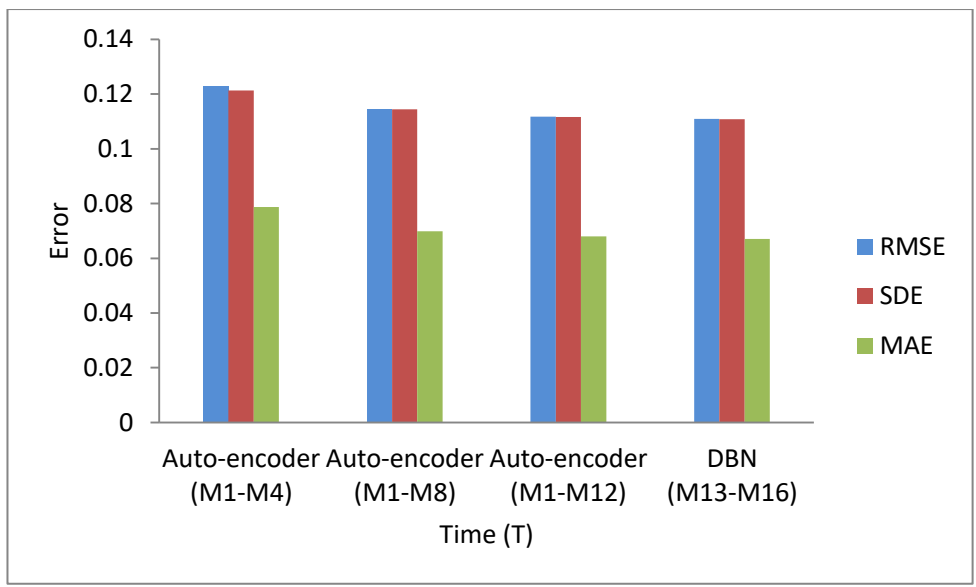

Figure 14: Performance of the proposed ATL-DNN technique on wind farm5

### 4.2. Actual and Predicted Power

Actual and predicted wind power are graphically shown in Figures 15-19. First two figures (Figure 15 and 16) show actual and predicted power related to wind farm1 and wind farm2, whereas Figure 17, Figure 18, and Figure 19 show actual and predicted power related to wind farm3, wind farm4, and wind farm5 datasets, respectively. All figures clearly represent that predicted power is close to actual power not only in case of wind farm2 (on which intra TL is employed) but also for the remaining four wind farms.

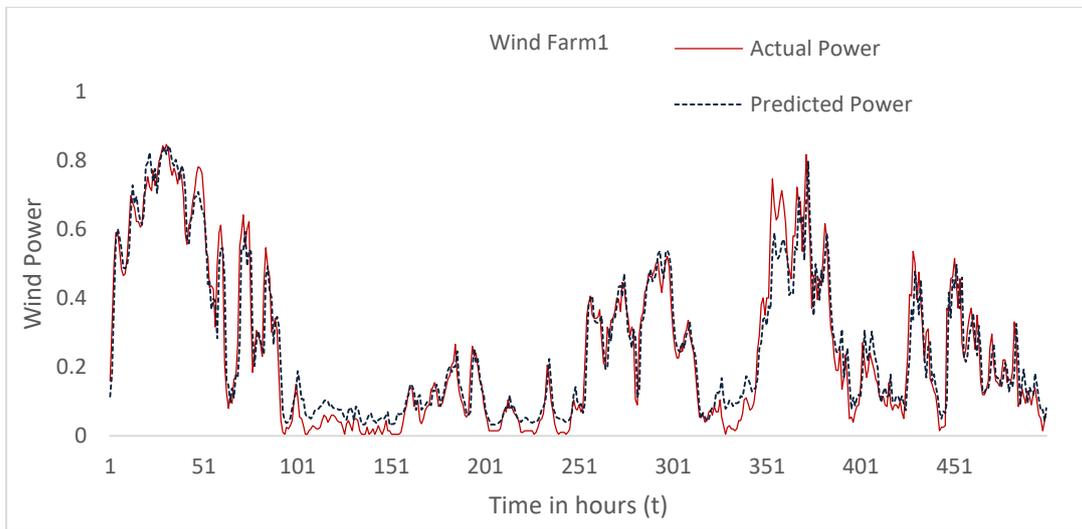

Figure 15: Predicted and actual power of the proposed ATL-DNN technique (wind farm1)

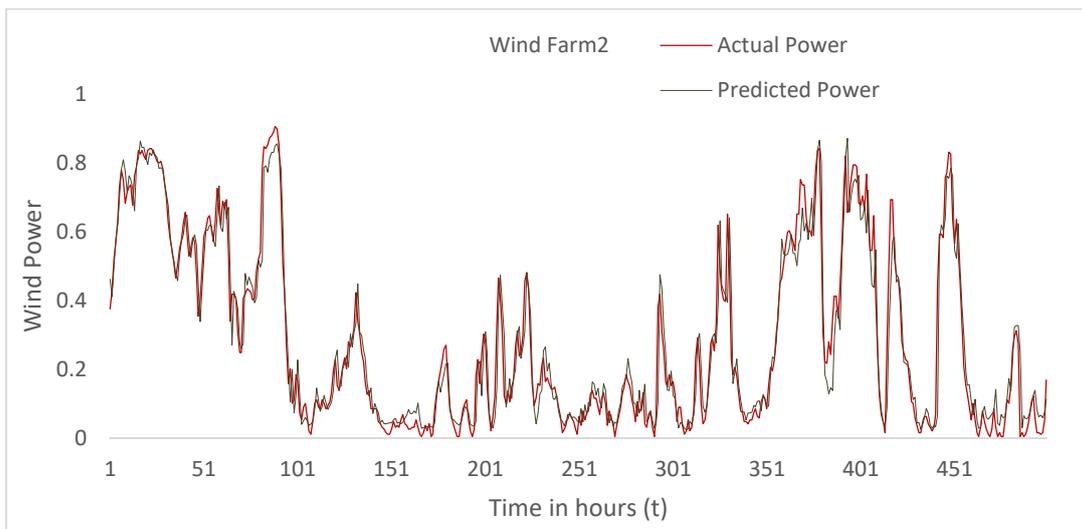

Figure 16: Predicted and actual power of the proposed ATL-DNN technique (wind farm2)
19

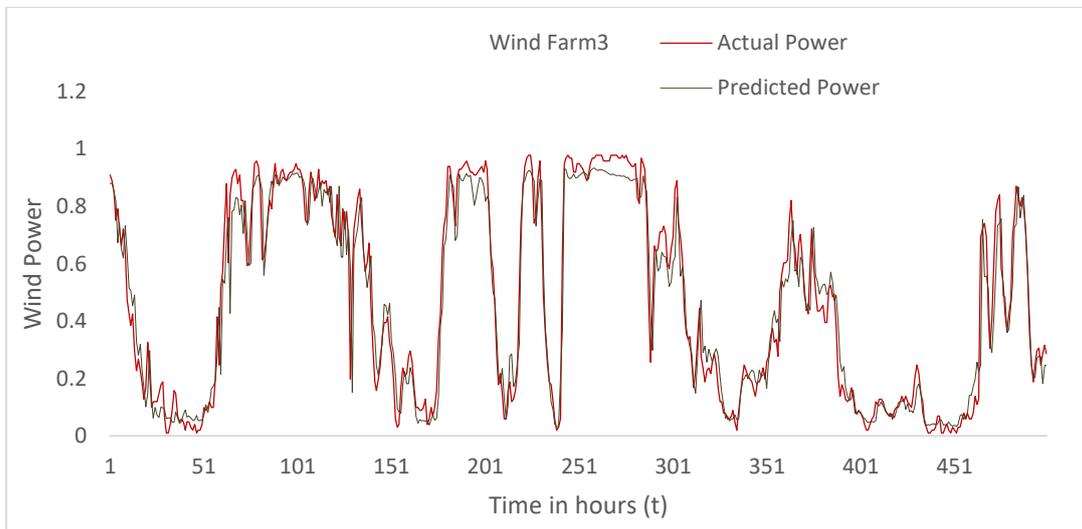

Figure 17: Predicted and actual power of the proposed ATL-DNN technique (wind farm3)

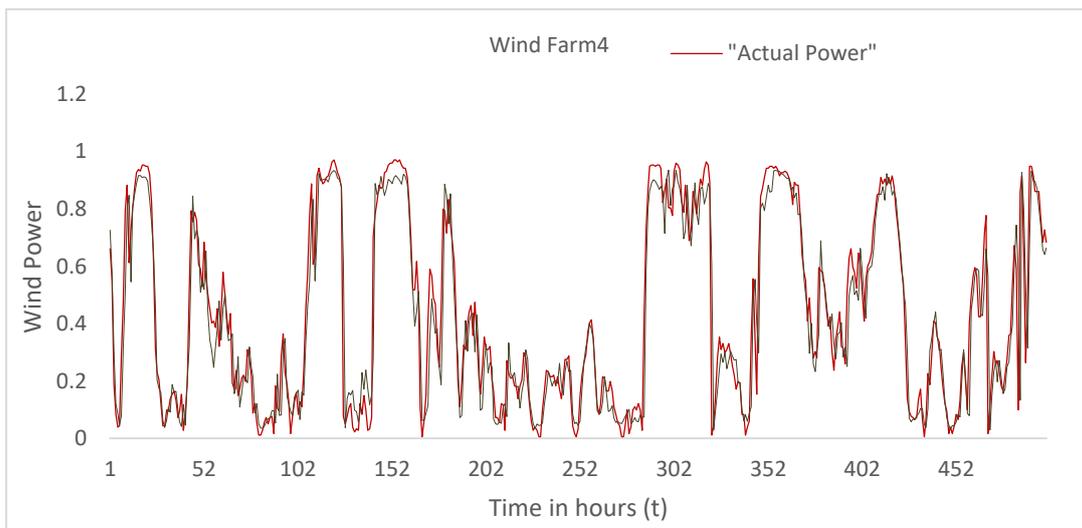

Figure 18: Predicted and actual power of the proposed ATL-DNN technique (wind farm4)

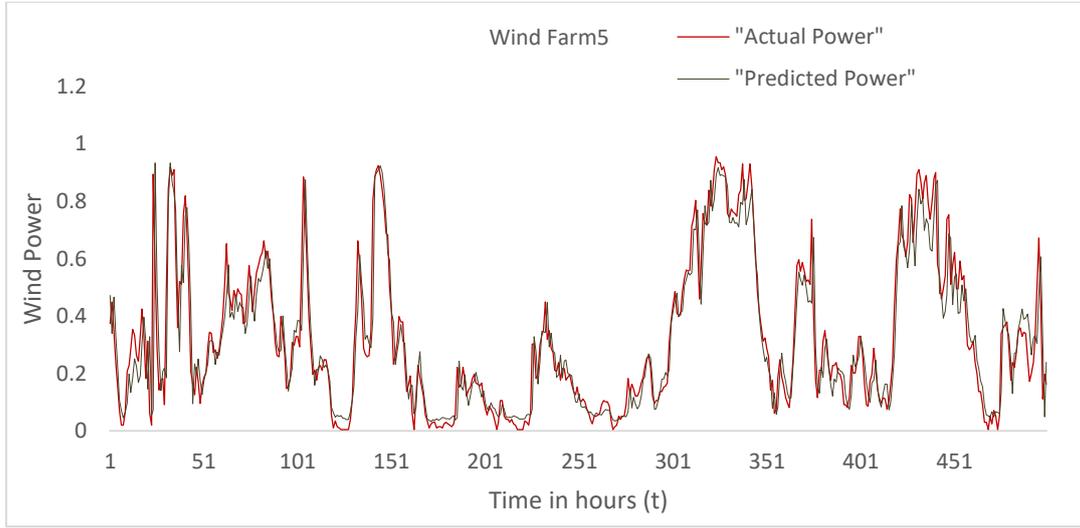

Figure 19: Predicted and actual power of the proposed ATL-DNN technique (wind farm5)

### 4.3. Performance of Meta-learner Using Data from Recently Trained Auto-encoder on 1-12 Months of Data

To evaluate the capabilities of the proposed ATL-DNN technique, only Predictions from the last auto-encoder (that is trained on 1-12 months of data) together with the original features are delivered as input to the meta-learner (DBN). Table 9 shows the performance when predictions from recently trained base-learner assist to increase the feature space for the meta-learner. Comparison of Table 9 with Table 6-8 shows that if predictions from three different base-learners (trained on a different sets of data) are concatenated along with original features then it will increase the performance of meta-learner, in comparison to the feature space formed by only concatenating the predictions from the last auto-encoder with the original features. The reason behind the increase in performance is that training of different base-learners on different data may extract robust and diverse features, which ultimately increases the generalization performance of the meta-learner.

Table 9: Performance of the proposed ATL-DNN technique using data obtained from a recently trained auto-encoder

|  | RMSE | MAE | SDE |
|---|---|---|---|
| **Wind farm1** | 0.0833 ± 0.0003 | 0.0590 ± 0.0003 | 0.07923 ± 0.0001 |
| **Wind farm2** | 0.1764 ± 0.0126 | 0.1352 ± 0.0088 | 0.1752 ± 0.0127 |
| **Wind farm3** | 0.1120 ± 0.0005 | 0.0731 ± 0.0004 | 0.1094 ± 0.0003 |
| **Wind farm4** | 0.1008 ± 0.0003 | 0.0677 ± 0.0003 | 0.1007 ± 0.0003 |
| **Wind farm5** | 0.1149 ± 0.0003 | 0.0719 ± 0.0004 | 0.1126 ± 0.0002 |

### 4.4. Inter TL for Transferring Knowledge from Region to Region and for Different Task

In order to evaluate the effectiveness of the proposed ATL-DNN technique, it is also tested on wind speed related data collected from a wind farm located in Pakistan. According to Betz's law, wind speed and power are directly related (equation 18).

$$p_{predict} = \frac{1}{2}\rho A V^3 c_p \qquad (18)$$

In equation 18, $\rho$ is the density of air, V is the wind speed, whereas is the power coefficient (which is unique to each wind turbine) and A is the Area of wind turbine. Therefore, for inter TL, pre-trained auto-encoder trained on a specific wind farm data from source domain, are adaptively trained after every two months on wind speed data set (collected from the wind farms of Pakistan). Table 10 shows the performance of meta-learner for ten independent runs. The low value of RMSE, MAE, and SDE shows that even with target domain data belonging to a different region and having different feature space, the proposed ATL-DNN technique performs well. Proposed adaptive ATL-DNN technique is thus suitable for transfer of knowledge across different regions. On the other hand, it also shows good results for transferring knowledge across different tasks (from wind speed to wind power predictions). Thus, the proposed ATL-DNN technique is not only robust to terrain related variations, but also exploits and transfer target function related information.

Table 10: Inter TL based performance of the proposed ATL-DNN technique wind speed prediction

| | WIND SPEED DATA | | |
|---|---|---|---|
| | RMSE | MAE | SDE |
| 1 | 0.0988 | 0.0772 | 0.0988 |
| 2 | 0.1456 | 0.1148 | 0.1334 |
| 3 | 0.1483 | 0.1242 | 0.1080 |
| 4 | 0.1406 | 0.1172 | 0.1080 |
| 5 | 0.1606 | 0.1356 | 0.1080 |
| 6 | 0.1557 | 0.1311 | 0.1080 |
| 7 | 0.0959 | 0.0751 | 0.0956 |
| 8 | 0.1540 | 0.1294 | 0.1080 |
| 9 | 0.1436 | 0.1200 | 0.1080 |
| 10 | 0.1560 | 0.13140 | 0.1080 |
| Mean | 0.1399 ± 0.0220 | 0.1156 ± 0.0207 | 0.1083 ± 0.0093 |

**4.5. Power Distribution Across Various Wind Farms**

Figure 20 shows that training and test sets of all the wind farms follow almost the same power distribution. As wind farm 1 and 2 have very similar power distribution, therefore, inter TL across wind farm2 (as source domain) and wind farm1 (as target domain) is quite successful in terms of evaluation measures. Similarly, prediction performance (using inter TL) for the other wind farms can be achieved by using those wind farms as the source domains that have almost a similar power distribution pattern.

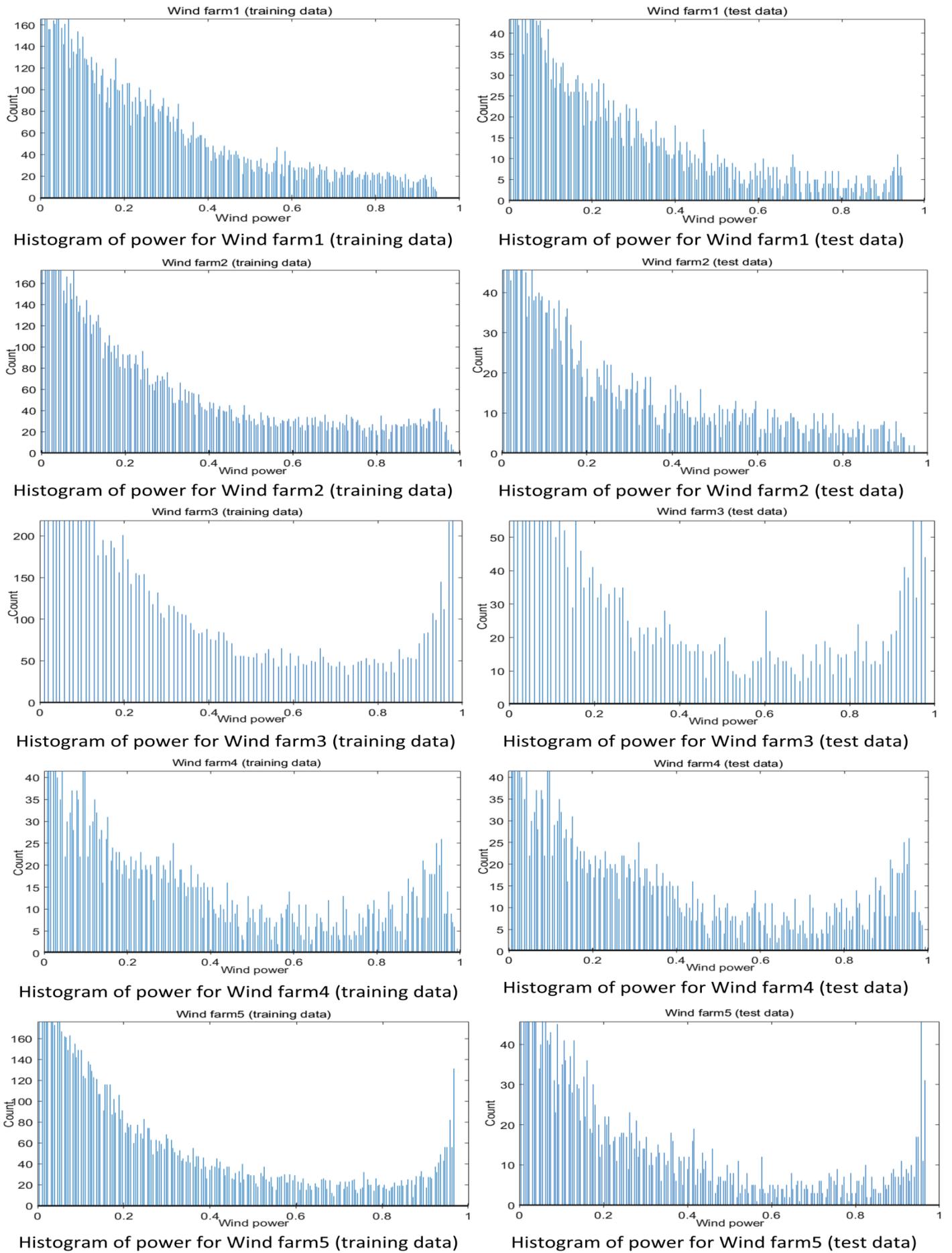

Figure 20: Distribution of power across wind farms

**4.6. Comparison of the Proposed ATL-DNN Technique with Existing Techniques**

Figure 21 shows the performance of the Proposed ATL-DNN technique in comparison to Grassi et al. [20], Amjady at al. [24], Zameer et al. [25], and Qureshi et al. [28] works. In case of the proposed ATL-DNN technique, it is noted that there is no need to train the base and meta-learners on the whole three years of wind farm1 data. During training, only pre-trained auto-encoder from wind farm1 is fine-tuned adaptively after every four months to form the three base-learners. And then, the final ensemble technique is directly evolved test data. Figure 21 shows that the performance of the proposed ATL-DNN technique in comparison to the existing techniques (Grassi et al. [20] , Amjady et al. [24], Zameer et al. [25], and Qureshi et al.[28] ) in terms of all the three evaluation matrices.

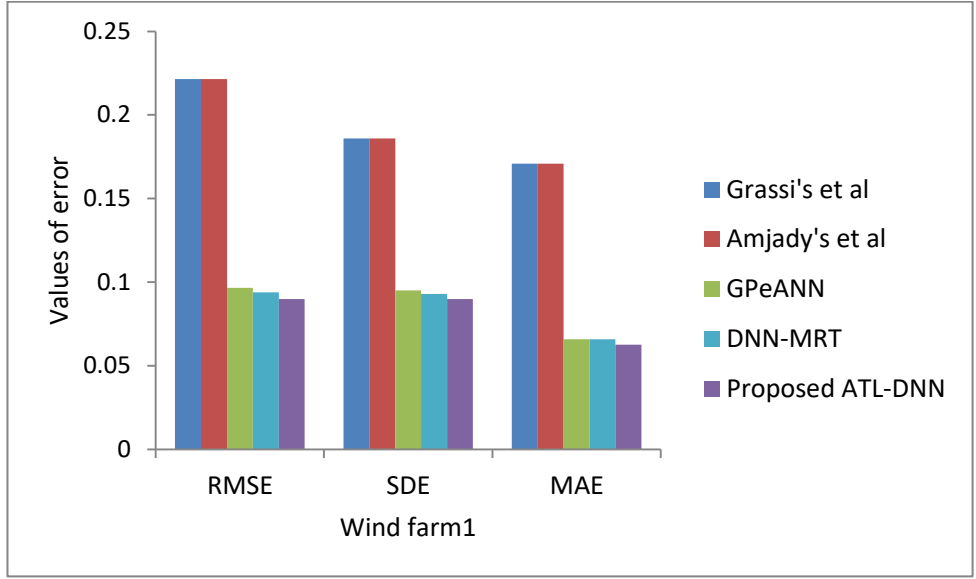

Figure 21: Performance comparison of the Proposed ATL-DNN technique against Grassi et al. [20],Amjady's et al. [24], Zameer et al. [25], and Qureshi et al.[28]

**4.7. Evaluation in Terms of Statistical Test**

In order to measure the relationship between actual and predicted wind power, Pearson Correlation Coefficient is used as a statistical performance measure. Pearson Correlation Coefficient was presented by Karl Pearson, which shows how two variables are linearly related to each other. Its range of values is between -1 and 1. Value of 1 shows that two variables are exactly linearly related to each other, whereas the value of zero shows no relationship and -1 depict a negative linear relationship. Let's suppose the actual power against wind farm is $\{x_1, x_2, x_3 .............x_p\}$ and the predicted wind power is $\{x_1', x_2', x_3'..........x_p'\}$ then, mathematically, Pearson Correlation Coefficient can be expressed as.

$$p_{corr} = \frac{\sum_{p=1}^{n}(x_i - x_m)(x_i' - x_m')}{\sqrt{\sum_{p=1}^{n}(x_i - x_m)^2 \sum_{i=1}^{n}(x_i' - x_m')^2}} \quad (19)$$

Where, $x_m$ and $x_m'$ are the mean values of actual and predicted wind power values, respectively.

Table 11 shows the Pearson Correlation values between the actual and predicted wind power for five wind farms. As value of Pearson correlation is approaching one for all of the five wind farms, which indicates a strong linear correlation between actual and predicted power.

Table 11: Pearson correlation for the five wind farm datasets

| Dataset | Pearson correlation |
|---|---|
| Wind Farm1 | 0.9438 |
| Wind Farm2 | 0.9308 |
| Wind Farm3 | 0.9441 |
| Wind Farm4 | 0.9437 |
| Wind Farm5 | 0.9243 |

## 5. Conclusion

An adaptive Deep Neural Network based technique is proposed, that uses the concept of inter and intra TL. Adaptive transfer learning not only helps in providing good weight initialization of the base-regressors, but also helps to better utilize the online data that is continuously being generated by wind farms. The proposed idea of "Adaptive Transfer Learning" is new and interesting, which is shown to experimentally make the ML technique suitable for real-time applications. Our proposed idea not only provides adaptability of the ML system, but it also provides the ability to the ML technique to transfer knowledge across different terrain based data. In addition, the experimental results show that the proposed ATL-DNN technique is also capable to share knowledge between different tasks (from wind power to speed prediction). It is shown that the concept of Inter and intra adaptive TL used in the proposed ATL-DNN technique is not only helpful in reducing the training time of individual base-learners, but also provides good average performance in terms of SDE, MAE, and RMSE for wind farms located in different regions.

**ACKNOWLEDGMENT**

This work is supported by the Higher Education Commission of Pakistan under the Indigenous Ph.D. Fellowship Program (PIN#213-54573-2EG2-097) and NRPU Research Grant (No. 20-3408/R&D/HEC/14/233). We also acknowledge Pakistan Institute of Engineering and Applied Sciences (PIEAS) for healthy research environment which leads to the research work presented in this article.

**Conflict of Interest:**

The authors declare that they have no conflict of interest.